\documentclass[11pt,a4paper]{article}
\usepackage[hyperref]{emnlp-ijcnlp-2019}
\usepackage{times}
\usepackage{latexsym}
\usepackage{graphicx}
\usepackage{amssymb}
\usepackage{subfig}
\usepackage{multirow}
\usepackage{xspace}
\usepackage{amsmath}
\usepackage{xcolor}
\usepackage{colortbl}

\usepackage{url}

\aclfinalcopy 

\usepackage{soul}
\setul{0.5ex}{0.4ex}
\newcommand{\tc}[2]{\setulcolor{#1}\ul{#2}\setulcolor{black}}

\newcommand{\tcA}[1]{\tc{blue}{#1}}
\newcommand{\tcB}[1]{\tc{orange}{#1}}

\newcommand{\tcD}[1]{\tc{green}{#1}}

\newcommand{\selectmod}{\rowcolor{purple!15}}
\newcommand{\mfdataset}{KGD\xspace}
\newcommand{\shortans}{OBQA-Short\xspace}
\newcommand{\obqafull}{OBQA-Full\xspace}
\newcommand{\gapqa}{GapQA\xspace}
\newcommand{\kermodel}{KER\xspace}

\newcommand{\qtof}{Question-to-Fact\xspace}
\newcommand{\ftoa}{Fact-to-Answer\xspace}
\newcommand{\qtoa}{Question-to-Answer(Fact)\xspace}

\newcommand{\ques}{\ensuremath{q}\xspace}
\newcommand{\qstem}{\ensuremath{q_{s}}\xspace}
\newcommand{\fact}{\ensuremath{f}\xspace}
\newcommand{\choices}{\ensuremath{c}\xspace}
\newcommand{\choice}{\ensuremath{c}\xspace}
\newcommand{\choicei}{\ensuremath{c_i}\xspace}
\newcommand{\spansym}{\ensuremath{s}\xspace}

\newcommand{\reln}{\ensuremath{r}\xspace}
\newcommand{\tuple}{\ensuremath{t}\xspace}
\newcommand{\kb}{\ensuremath{K}\xspace}
\newcommand{\kbj}{\ensuremath{k_j}\xspace}

\newcommand{\qtoks}{\ensuremath{q_{m}}\xspace}
\newcommand{\ftoks}{\ensuremath{\fact_m}\xspace}
\newcommand{\hdim}{\ensuremath{h}\xspace}

\newcommand{\pred}[1]{\ensuremath{\Hat{#1}}\xspace}
\newcommand{\gold}[1]{\ensuremath{\Bar{#1}}\xspace}

\newcommand{\enc}[1]{\ensuremath{\mathcal{E}_{#1}}\xspace}
\newcommand{\attm}[2]{\ensuremath{\mathcal{A}_{{#1}, {#2}}}\xspace}

\newcommand{\normatt}[2]{\ensuremath{\mathcal{V}_{#1}( #2)}\xspace}
\newcommand{\softmax}{\mathrm{softmax}}
\newcommand{\size}[2]{\mathbb{R}^{#1\times#2}}
\newcommand{\wtrep}[2]{\mathcal{S}_{#1}(#2)}
\newcommand{\score}{\mathrm{score}}
\newcommand{\mlp}{\ensuremath{\mathrm{FF}}\xspace}
\newcommand{\compsn}[2]{\ensuremath{\bigotimes(#1,#2)}}
\newcommand{\relnrep}{\ensuremath{\mathcal{R}}}
\newcommand{\avg}{\mathrm{avg}}

\newcommand\T{\rule{0pt}{2.6ex}}       
\newcommand\B{\rule[-1.2ex]{0pt}{0pt}} 

\title{What's Missing: A Knowledge Gap Guided Approach\\
for Multi-hop Question Answering}

\author{Tushar Khot \and Ashish Sabharwal \and Peter Clark\\
  Allen Institute for Artificial Intelligence,
  Seattle, WA, U.S.A.\\
  {\tt \small \{tushark,ashishs,peterc\}@allenai.org}
  }

\date{}

\begin{document}
\maketitle

\begin{abstract}
Multi-hop textual question answering requires combining information from multiple sentences. We focus on a natural setting where, unlike typical reading comprehension, only \emph{partial information} is provided with each question. The model must retrieve and use additional knowledge to correctly answer the question. To tackle this challenge, we develop a novel approach that explicitly identifies the \emph{knowledge gap} between a key span in the provided knowledge and the answer choices. The model, GapQA, learns to fill this gap by determining the relationship between the span and an answer choice, based on retrieved knowledge targeting this gap. We propose jointly training a model to simultaneously fill this knowledge gap and compose it with the provided partial knowledge. On the OpenBookQA dataset, given partial knowledge, explicitly identifying what's missing substantially outperforms previous approaches.
\end{abstract}

\section{Introduction}

Reading Comprehension datasets~\cite{Richardson2013-mctest-dataset,Rajpurkar2016-squad,joshi-EtAl:2017:Trivia-qa} have gained interest as benchmarks to evaluate a system's ability to understand a document via question answering (QA). Since many of these early datasets only required a system to understand a single sentence, new datasets were specifically designed to focus on the problem of multi-hop QA, i.e., reasoning across sentences~\cite{MultiRCKhashabi2018,wikihop,hotpotqa}.

While this led to improved language understanding, the tasks still assume that a system is provided with \emph{all} knowledge necessary to answer the question. In practice, however, we often only have access to \emph{partial knowledge} when dealing with such multi-hop questions, and must retrieve additional facts (the knowledge ``gaps'') based on the question and the provided knowledge. Our goal is to identify such gaps and fill them using an external knowledge source.

\begin{figure}[t]
\centering
\fbox{
\begin{minipage}{0.44\textwidth}
\small
\textbf{Question:}\\
\emph{Which of these would \tcA{let the most heat travel through}?}\\
A) a new pair of jeans. \\
B) \tcD{a steel spoon in a cafeteria}. \\
C) a cotton candy at a store. \\
D) a calvin klein cotton hat. \\
\noindent
\textbf{Core Fact:} \\
\tcB{Metal} \tcA{lets heat travel through}. \\
\\
\textbf{Knowledge Gap} (similar gaps for other choices):\\
\tcD{steel spoon in a cafeteria} \underline{\hspace{1cm}} \tcB{metal}.\\
\\
\textbf{Filled Gap} (relation identified using KB):\\
\tcD{steel spoon in a cafeteria} \emph{is made of} \tcB{metal}.
\end{minipage}
}
\caption{\label{figure:dataset-example}  A sample OpenBookQA question, the identified knowledge gap based on partial information in the core fact, and relation (\emph{is made of}) identified from a KB to fill that gap.
}
\end{figure}

The recently introduced challenge of \emph{open book} question answering~\cite{openbookqa} highlights this phenomenon. The questions in the corresponding dataset, OpenBookQA,
are derived from a science fact in an ``open book'' of about 1300 facts. To answer these questions, a system must not only identify a relevant ``core'' science fact from this small book, but then also retrieve additional common knowledge from large external sources in order to successfully apply this core fact to the question. Consider the example in Figure~\ref{figure:dataset-example}. The core science fact \emph{metal lets heat to travel through} points to \emph{metal} as the correct answer, but it is not one of the 4 answer choices. Given this core fact (the ``partial knowledge''), a system must still use broad external knowledge to fill the remaining gap, that is, identify which answer choice \emph{contains} or \emph{is made of} metal. 

This work focuses on \emph{QA under partial knowledge}. This turns out to be a surprisingly challenging task in itself; indeed, the partial knowledge models of \citet{openbookqa} achieve a score of only 55\% on OpenBookQA, far from human performance of 91\%. Since this and several recent multi-hop datasets use the multiple-choice setting~\cite{wikihop,MultiRCKhashabi2018,Lai2017RACELR}, we assume access to potential answers to a question. While our current model relies on this for a direct application to span-prediction based RC datasets, the idea of identifying knowledge gaps can be used to create novel RC specific models.

We demonstrate that an intuitive approach leads to a strong model:
first identify the knowledge gap and then fill this gap, i.e., identify the missing relation using external knowledge. We primarily focus on the OpenBookQA dataset since it is the only dataset currently available that provides partial context. However, we believe such an approach is also applicable to the broader setting of multi-hop RC datasets, where the system could start reasoning with one sentence and fill remaining gap(s) using sentences from other passages.

Our model operates in two steps. First, it predicts a key span in the core fact (``metal'' in the above example). Second, it answers the question by identifying the relationship between the key span and answer choices, i.e., by \emph{filling} the knowledge gap. This second step can be broken down further: (a) retrieve relevant knowledge from resources such as ConceptNet~\cite{Speer2017Conceptnet55} and large-scale text corpora~\cite{ARCClark2018}; (c) based on this, predict potential relations between the key span and an answer choice; and (d) compose the core fact with this filled gap.

We collect labels for knowledge gaps on $\sim$30\% of the training questions, and train two modules capturing the two main steps above. The first exploits an existing RC model and large-scale dataset to train a span-prediction model.
The second uses multi-task learning to train a separate QA model to jointly predict the relation representing the gap, as well as the final answer. For questions without labelled knowledge gaps, the QA model is trained based solely on the predicted answer.

Our model outperforms the previous state-of-the-art partial knowledge models by 6.5\% (64.41 vs 57.93) on a targeted subset of OpenBookQA amenable to gap-based reasoning. Even without missing fact annotations, our model with a simple heuristic to identify missing gaps still outperforms previous models by 3.4\% (61.38 vs.~57.93). It also generalizes to questions that were not its target, with 3.6\% improvement (59.40 vs.~55.84) on the full OpenBookQA set.

Overall, the contributions of this work are: (1) an analysis and dataset\footnote{\label{footnote:code-and-data}The code and associated dataset are available at https://github.com/allenai/missing-fact.} of knowledge gaps for QA under partial knowledge;
(2) a novel two-step approach of first identifying and then filling knowledge gaps for multi-hop QA;
(3) a model$^{\textrm{\ref{footnote:code-and-data}}}$ that simultaneously learns to fill a knowledge gap using retrieved external knowledge and compose it with partial knowledge; and
(4) new state-of-the-art results on QA with partial knowledge (+6.5\% using annotations on only 30\% of the questions).

\section{Related Work}

\paragraph{Text-Based QA.}
Reading Comprehension (RC) datasets probe language understanding via question answering. While several RC datasets~\cite{Rajpurkar2016-squad,Trischler2017-rc-newsqa,joshi-EtAl:2017:Trivia-qa} can be addressed with single sentence understanding, newer datasets~\cite{babi-Weston-15,wikihop,MultiRCKhashabi2018,hotpotqa} specifically target multi-hop reasoning. In both cases, all relevant information, barring some linguistic knowledge, is provided or the questions are unanswerable~\cite{squad2}. This allows using an attention-based approach of indirectly combining information~\cite{corefgru,gcn_entity,mhqa_grn}.

On the other hand, open domain question answering datasets~\cite{clark2016combining,ARCClark2018} come with no context, and require first retrieving relevant knowledge before reasoning with it. Retrieving this knowledge from noisy textual corpora, while simultaneously solving the reasoning problem, can be challenging, especially when questions require multiple facts.
This results in simple approaches (e.g. word-overlap/PMI-based approaches), that do not heavily rely on the retrieval quality, being competitive with other complex reasoning methods that assume clean knowledge~\cite{clark2016combining,jansen2017framing,2016angeli-naturalli}. To mitigate this issue, semi-structured tables~\cite{tableilp2016,Jansen2018WorldTreeAC} have been manually authored targeting a subset of these questions. However, these tables are expensive to create and these questions often need multiple hops (sometimes up to 16~\cite{Jansen2018WorldTreeAC}), making reasoning much more complex.

OpenBookQA dataset~\cite{openbookqa} was proposed to limit the retrieval problem by providing a set of $\sim$1300 facts as an `open book` for the system to use. Every question is based on one of the core facts, and in addition requires basic external knowledge such as hypernymy, definition, and causality. We focus on the task of question answering under partial context, where the core fact for each question is available to the system.

\begin{figure*}
      \begin{minipage}[b]{0.48\linewidth}
      \centering
        \includegraphics[width=0.95\linewidth]{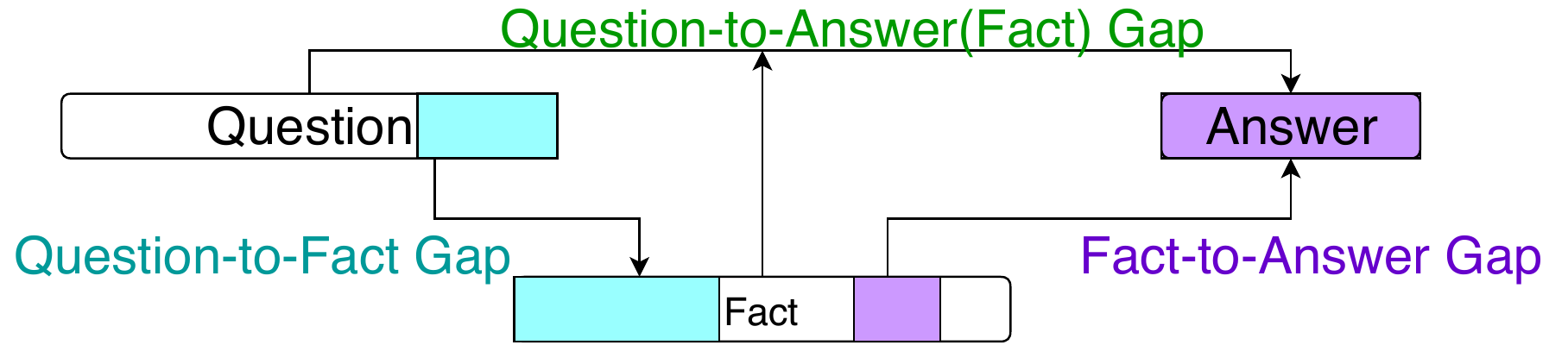}
        \caption{High-level overview of the kinds of knowledge gaps, assuming partial knowledge from the fact. In subsequent figures, the knowledge gap is indicated using \hl{highlighted} text.}
        \label{fig:kg_high}
    \end{minipage}
    \hspace{1ex}
    \begin{minipage}[b]{0.48\linewidth}
    \centering
        \includegraphics[width=\linewidth]{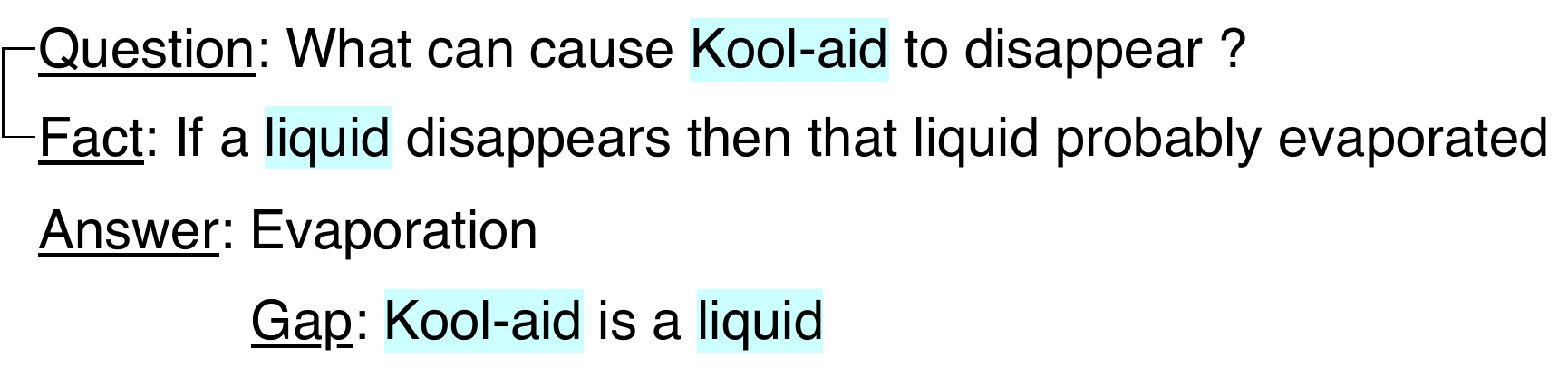}
        \caption{Knowledge gap between the \textbf{question} (Kool-aid) and the \textbf{fact} (liquid). To apply the fact about liquids to the question, we need to know ``Kool-aid is a liquid''.}
        \label{fig:kg_qtof}
    \end{minipage}
    \begin{minipage}[b]{0.48\linewidth}
    \centering
        \includegraphics[width=\linewidth]{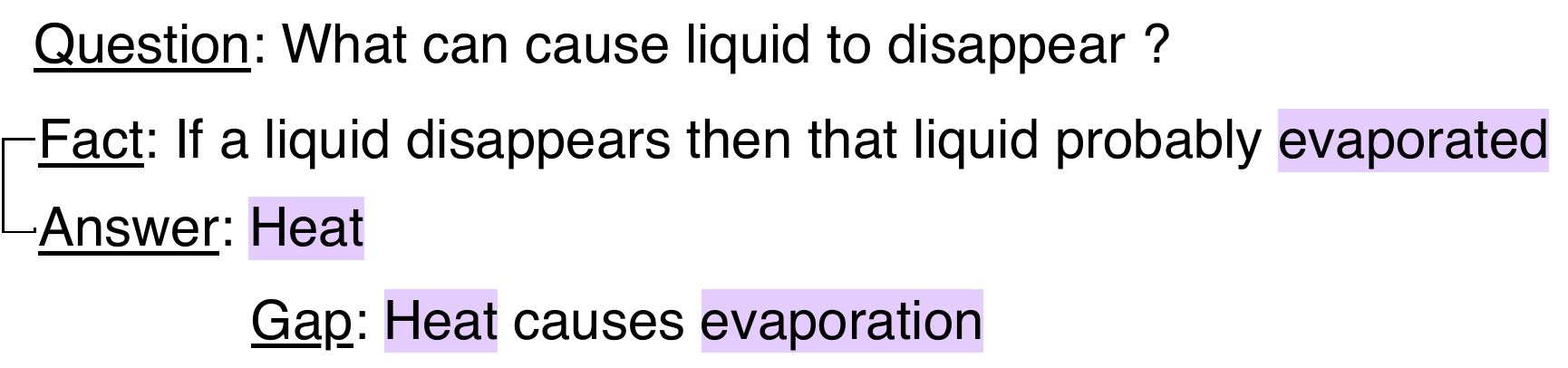}
        \caption{Knowledge gap between the \textbf{fact} (evaporated) and the \textbf{answer} (Heat). While it is clear how to apply the knowledge, we need to know that ``Heat causes evaporation'' to identify the right answer.}
        \label{fig:kg_ftoa}
    \end{minipage}
    \hspace{1ex}
    \begin{minipage}[b]{0.48\linewidth}
    \centering
        \includegraphics[width=0.9\linewidth]{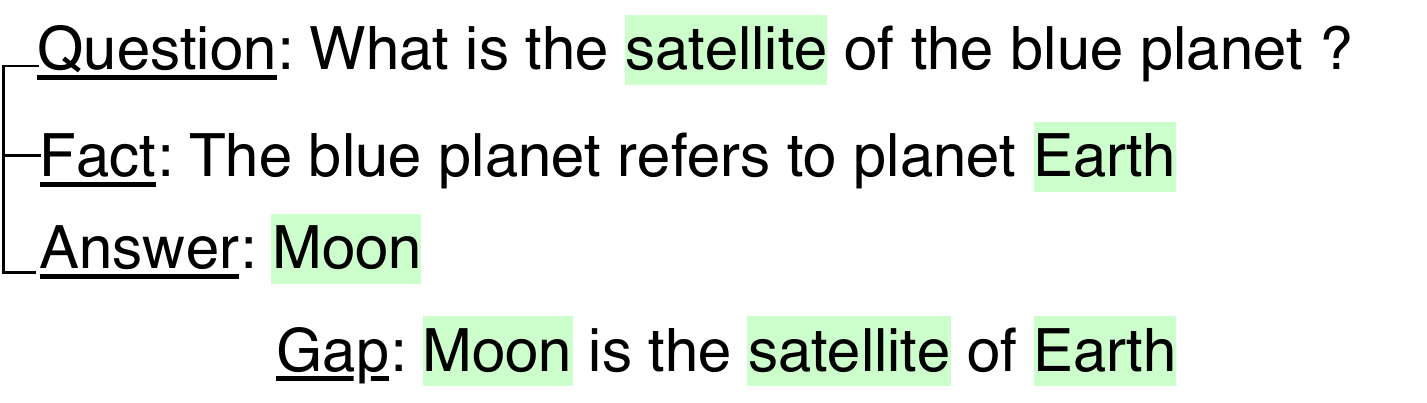}
        \caption{Knowledge gap between the \textbf{question} and the \textbf{answer} using the \textbf{fact}. For some complex questions, the fact clarifies certain concepts in the question, (e.g., ``blue planet''), leading to a reformulation of the question based on the fact (e.g., ``What is the satellite of Earth?'') which is captured by this gap.}
        \label{fig:kg_qtoa}
    \end{minipage}
\end{figure*}

\paragraph{Knowledge-Based QA.}
Another line of research is answering questions~\cite{simplequestions,wikitable,webquestions} over a structured knowledge base (KB) such as Freebase~\cite{bollacker2008freebase}. Depending on the task, systems map questions to a KB query with varying complexity: from complex semantic parses~\cite{Krishnamurthy2017NeuralSP} to simple relational lookup~\cite{Petrochuk2018SimpleQuestionsNS}. Our sub-task of filling the knowledge gap can be viewed as KB QA task with knowledge present in a KB or expected to be inferred from text. 

Some RC systems~\cite{Mihaylov2018EnhanceCS,kadlec-EtAl:2016:P16-1} and Textual Entailment (TE) models~\cite{Weissenborn2018DynamicIO,Inkpen2018NeuralNL} incorporate external KBs to provide additional context to the model for better language understanding. However, we take a different approach of using this background knowledge in an explicit inference step (i.e. hop) as part of a multi-hop QA model.

\section{Knowledge Gaps}

We now take a deeper look at categorizing knowledge gaps into various classes. While grounded in OpenBookQA, this categorization is relevant for other multi-hop question sets as well.\footnote{As mentioned earlier, in the RC setting, the first relevant sentence read by the system can be viewed as the core fact.} We will then discuss how to effectively annotate such gaps.

\subsection{Understanding Gaps: Categorization}

We analyzed the additional facts needed for answering 75 OpenBookQA questions. These facts naturally fall
into three classes, based on the knowledge gap they are trying to fill: (1) \qtof, (2) \ftoa, and (3) \qtoa.  Figure~\ref{fig:kg_high} shows a high-level overview, with simplified examples of each class of knowledge gap in Figures~\ref{fig:kg_qtof},~\ref{fig:kg_ftoa}, and~\ref{fig:kg_qtoa}.

\paragraph{\qtof Gap.}
This gap exists between concepts in the question and the core fact. For example, in Figure~\ref{fig:kg_qtof}, the knowledge that ``Kool-aid'' is a liquid is needed to even recognize that the fact is relevant.

\paragraph{\ftoa Gap.}
This gap captures the relationship between concepts in the core fact and the answer choices. For example, in Figure~\ref{fig:kg_ftoa}, the knowledge ``Heat causes evaporation'' is needed to relate ``evaporated'' in the fact to the correct answer ``heat''. Note that it is often possible to find relations connecting the fact to even incorrect answer choices. For example,  ``rainfall'' could be connected to the fact using ``evaporation leads to rainfall''. Thus, identifying the correct relation and knowing if it can be composed with the core fact is critical, i.e., ``evaporation causes liquid to disappear'' and ``evaporation leads to rainfall'' do not imply that ``rainfall causes liquid to disappear''. 

\paragraph{\qtoa Gap.}
Finally, some questions need additional knowledge to connect concepts in the question to the answer, based on the core fact. For example, composition questions (Figure~\ref{fig:kg_qtoa}) use the provided fact to replace parts of the original question with words from the fact.

Notably \qtof and \ftoa gaps are more common in OpenBookQA (44\% and 86\% respectively\footnote{Some questions have both of these classes of gaps.}), while the \qtoa gap is very rare ($<$20\%). While all three gap classes pose important problems, we focus on \ftoa gap and assume that the core fact is provided. This is still a challenging problem as one must not only identify and fill the gap, but also learn to compose this filled gap with the input fact.
\subsection{Annotating Gaps: Data Collection}

\begin{table*}[t]
    \setlength{\doublerulesep}{\arrayrulewidth}
    \small
    \centering
    \begin{tabular}{p{4.2cm}|p{3.5cm}|p{1.3cm}|p{1.5cm}|p{2.8cm}}
    \\   
     Question & Fact & Span & Relation & Gap\\
    \hline\hline
   \textbf{Q:} A light bulb turns on when it receives energy from \textbf{A:} gasoline & a light bulb converts \textit{electrical energy} into light energy when it is turned on & electrical energy & provides$^{-1}$, enables$^{-1}$  & (gasoline, provides, electrical energy)\\
      \textbf{Q:} What makes the best wiring? \textbf{A:} Tungsten & wiring requires an \textit{electrical conductor} & electrical conductor & isa$^{-1}$, madeof & (Tungsten, is an, electrical conductor)
    \\ \hline
    \end{tabular}
    \caption{Examples from \mfdataset\ dataset. Note that the knowledge gap is captured in the form of (Span, Relation, Answer) but not explicitly annotated. $^{-1}$ is used to indicate that the argument order should be flipped.}
    \label{tab:kgd_egs}
\end{table*}

Due to space constraints, details of our crowdsourcing process for annotating knowledge gaps, including the motivation behind various design choices as well as several examples, are deferred to the Appendix (Section~\ref{appendix:data-collection}). Here we briefly summarize the final crowdsourcing design.

Our early pilots revealed that straightforward approaches to annotate knowledge gaps for all OpenBookQA questions lead to noisy labels. To address this, we (a) identified a \textbf{subset of questions} suitable for this annotation task and (b) split \ftoa gap annotation into \textbf{two steps}: key term identification and relation identification.

\textbf{Question Subset.} First, we identified valid question-fact pairs where the fact supports the correct answer (verified via crowdsourcing) but does not trivially lead to the answer (fact only overlaps with the correct answer). Second, we noticed that the \ftoa\ gaps were much noisier for longer answer options, where you could write multiple knowledge gaps or a single complex knowledge gap. So we created \emph{\shortans}, the subset of OpenBookQA where answer choices have at most two non-stopword tokens. This contains over 50\% of the original questions and is also the target set of our approach.

\textbf{Two-step Gap Identification.} Starting with the above pairs of questions with valid partial knowledge, the second task is to author facts that close the \ftoa \textbf{knowledge gap}. Again, initial iterations of the task resulted in poor quality, with workers often writing noisy facts that re-stated part of the provided fact or directly connect the question to the answer (skipping over the provided fact\footnote{This was also noticed by the original authors of OpenBookQA dataset~\cite{openbookqa}.}). We noticed that the core fact often contains a key span that hints at the final answer. So we broke the task into two steps: (1) identify key terms (preferably a span) in the core fact that could answer the question, and (2) identify one or more relations\footnote{Workers preferably chose from a selected list of nine most common relations: \{causes, definedAs, enables, isa, located in, made of, part of, provides, synonym of\} and their inverses (except synonymy). These relations have also been found to be useful by prior approaches for science QA~\cite{clark2014automatic,tableilp2016,Jansen2016Explanation,Jansen2018WorldTreeAC}.} that hold between the key terms and the correct answer choice but not the incorrect choices. Table~\ref{tab:kgd_egs} shows example annotations of the gaps obtained through this process.

\paragraph{Knowledge Gap Dataset: \mfdataset}\mbox{}

\begin{table}[t]
    \setlength{\doublerulesep}{\arrayrulewidth}
    \small
    \centering
    \begin{tabular}{l@{\hskip 1cm}ccc}
     & Train & Dev & Test\\
    \hline\hline
    \T  Total \#questions & 1151&117&121 \\
        Total \#question-facts & 1531&157&165 \\
        Avg. \# spans & 1.43&1.46&1.45 \\
        Avg. \# relations & 3.31&2.45&2.45
    \end{tabular}
    \caption{Statistics of the train/dev/test split of the \mfdataset dataset. The \#question-fact pairs is higher than \#questions as some questions may be supported by multiple facts. The average statistic computes the average number of unique spans and relations per question-fact pair.}
    \label{tab:data_stats}
\end{table}

Our Knowledge Gap Dataset (\mfdataset) contains key span and relation label annotations to capture knowledge gaps. To reduce noise, we only use knowledge gap annotations where at least two of three workers found a contiguous span from the core fact and a relation from our list. The final dataset contains examples of the form \{question, fact, spans, relations\}, where each span is a substring of the input fact, and relations are the set of valid relations between the span and the correct answer (examples in Table~\ref{tab:kgd_egs} and stats in Table~\ref{tab:data_stats}).

\section{Knowledge-Gap Guided QA: \gapqa}

We first introduce the notation used to describe our QA system. For each question \ques and fact \fact, the selected span is given by \spansym and the set of valid relations between this span and the correct choice is given by \reln. Borrowing notation from OpenBookQA, we refer to the question without the answer choices \choices as the stem \qstem, i.e., \ques = \qstem \choices. We use $\Hat{\spansym}$ to indicate the predicted span and $\Hat{\reln}$ for the predicted relations. We use \qtoks and \ftoks to represent the tokens in the question stem and fact respectively. Following the Turk task, our model first identifies the key span from the fact and then identifies the relation using retrieved knowledge.

\subsection{Key Span Identification Model}
Since the span selected from the fact often tends to be the answer to the question (c.f. Table~\ref{tab:kgd_egs}), we can use a reading comprehension model to identify this span. The fact serves as the input passage and the question stem as the input question to the reading comprehension model. We used the Bi-Directional Attention Flow (BiDAF) model~\cite{bidaf}, an attention-based span prediction model designed for the SQuAD RC dataset~\cite{Rajpurkar2016-squad}. We refer the reader to the original paper for details about the model. 

\subsection{Knowledge Retrieval Module}
Given the predicted span, we retrieve knowledge from two sources: triples from ConceptNet~\cite{Speer2017Conceptnet55} and sentences from ARC corpus~\cite{ARCClark2018}. ConceptNet contain (subject, relation, object) triples with relations such as \emph{/r/IsA}, \emph{/r/PartOf} that closely align with the relations in our gaps. Since ConceptNet can be incomplete or vague (e.g. \emph{/r/RelatedTo} relation), we also use the ARC corpus of 14M science-relevant sentences to improve our recall.

\paragraph{Tuple Search.} To find relevant tuples connecting the predicted span $\pred{\spansym}$ to the answer choice \choicei, we select tuples where at least one token\footnote{We use lower-cased, stemmed, non-stopword tokens.} in the subject matches $\pred{\spansym}$ and at least one token in the object matches \choicei (or vice versa). We then score each tuple \tuple using the Jaccard score\footnote{score(\tuple) = jacc(tokens(\tuple), tokens(\pred{\spansym} + \choicei)) where jacc(w1, w2) = $\frac{w1 \cap w2}{w1 \cup w2}$} and pick the top $k$ tuples for each \choicei ($k=5$ in our experiments).

\paragraph{Text Search.} To find the relevant sentences for $\pred{\spansym}$ and \choicei, we used ElasticSearch\footnote{https://www.elastic.co/products/elasticsearch} with the query: $\pred{\spansym}$ + \choicei (refer to Appendix~\ref{app:text_retrieval} for more details). Similar to ConceptNet, we pick top 5 sentences for each answer choice. To ensure a consistent formatting of all knowledge sources, we convert the tuples into sentences using few hand-defined rules(described in Appendix~\ref{sec:conceptnet}). Finally all the retrieved sentences are combined to produce the input KB for the model, \kb.

\subsection{Question Answering Model}

The question answering model takes as input the question \qstem, answer choices \choices, fact \fact, predicted span, \pred{\spansym} and retrieved knowledge \kb. We use 300-dimensional 840B GloVe embeddings~\cite{pennington2014glove} to embed each word in the inputs. We use a Bi-LSTM with 100-dimensional hidden states to compute the contextual encodings for each string, e.g., $\enc{\fact} \in \size{\ftoks}{\hdim}$. The question answering model selects the right answer using two components: (1) \textbf{Fact Relevance} module (2) \textbf{Relation Prediction} module.

\paragraph{Fact Relevance.}
This module is motivated by the intuition that a relevant fact will often capture a relation between concepts that align with the question and the correct answer (the cyan and magenta regions in Figure~\ref{fig:kg_high}). 
To deal with the gaps between these concepts, this module relies purely on word embeddings while the next module will focus on using external knowledge. 

We compute a question-weighted and answer-weighted representation of the fact to capture the part of the fact that links to the question and answer respectively. We compose these fact-based representations to then identify how well the answer choice is supported by the fact. 

To calculate the question-weighted fact representation, we first identify facts words with a high similarity to some question word ($\normatt{\qstem}{\fact}$)  using the attention weights: $\attm{\qstem}{\fact} = \enc{\qstem} \cdot \enc{\fact} \in \size{\qtoks}{\ftoks}$
\begin{align*}
    \normatt{\qstem}{\fact} = \softmax_{\ftoks}\left(\max_{\qtoks} \attm{\qstem}{\fact}\right) \in \size{1}{\ftoks}
\end{align*}
The final attention weights are similar to the Query-to-Context attention weights in BiDAF. The final question-weighted representation is:
\begin{align}
    \wtrep{\qstem}{\fact} = \normatt{\qstem}{\fact} \cdot \enc{\fact} \in \size{1}{\hdim}
    \label{eqn:wtrep}
\end{align}
We similarly compute the choice-weighted representation of fact as $\wtrep{\choicei}{\fact}$. We compose these two representations by averaging\footnote{We found this simple composition function performed better than other composition operations.} these two vectors $\wtrep{\qstem\choicei}{\fact} =  (\wtrep{\qstem}{\fact} +  \wtrep{\choicei}{\fact})/2$. We finally score the answer choice by comparing this representation with the aggregate fact representation, obtained by averaging too, as:
\begin{align*}
    \score_f(\choicei) = \mlp\left(\compsn{\wtrep{\qstem\choicei}{\fact}}{\avg(\enc{\fact}}\right)
\end{align*}
where $\compsn{x}{y} = [x-y; x*y] \in \size{1}{2\hdim}$ and $\mlp$ is a feedforward neural network that outputs a scalar score for each answer choice. 

\paragraph{Filling the Gap: Relation Prediction.}
The relation prediction module uses the retrieved knowledge to focus on the \ftoa gap by first predicting the relation between \pred{\spansym} and \choicei and then compose it with the fact to score the choice. We first compute the span and choice weighted representation($\size{1}{\hdim}$) for each sentence \kbj in \kb using the same operations as above: 
\begin{align*}
    \wtrep{\pred{\spansym}}{\kbj} = \normatt{\pred{\spansym}}{\kbj} \cdot \enc{\kbj} ~;~
    \wtrep{\choicei}{\kbj} = \normatt{\choicei}{\kbj} \cdot \enc{\kbj}
\end{align*}
These representations capture the contextual embeddings of the words in the \kbj that most closely resemble words in \pred{\spansym} and \choicei respectively. We predict the kb-based relation between them based on the composition of these representations :
\begin{align*}
    \relnrep_j(\pred{\spansym},\choicei) = \mlp\left(\compsn{\wtrep{\pred{\spansym}}{\kbj}}{\wtrep{\choicei}{\kbj}}\right) \in \size{1}{h}
\end{align*}
We pool the relation representations from all the KB facts into a single prediction by averaging, i.e. $\relnrep(\pred{\spansym},\choicei) = \avg_j\  \relnrep_j(\pred{\spansym},\choicei)$.

\paragraph{Relation Prediction Score.}
We first identify the potential relations that can be composed with the fact, given the question, e.g., in Figure~\ref{figure:dataset-example}, we can compose the fact with (steel spoon; \emph{made of}; metal) relation but not (metal; \emph{made of}; ions). We compose an aggregate representation of the question and fact encoding to capture this information:
\begin{align*}
    \mathcal{D}(\qstem, \fact) = \compsn{\max_{\qtoks} \enc{\qstem}}{\max_{\ftoks} \enc{\fact}} \in \size{1}{2h}
\end{align*}
We finally score the answer choice based on this representation and the relation representation:
\begin{align*}
    \score_r(\choicei) = \mlp\left([\mathcal{D}(\qstem, \fact);  \relnrep(\pred{\spansym},\choicei)]\right)
\end{align*}

\begin{figure}
    \centering
    \includegraphics[scale=0.36]{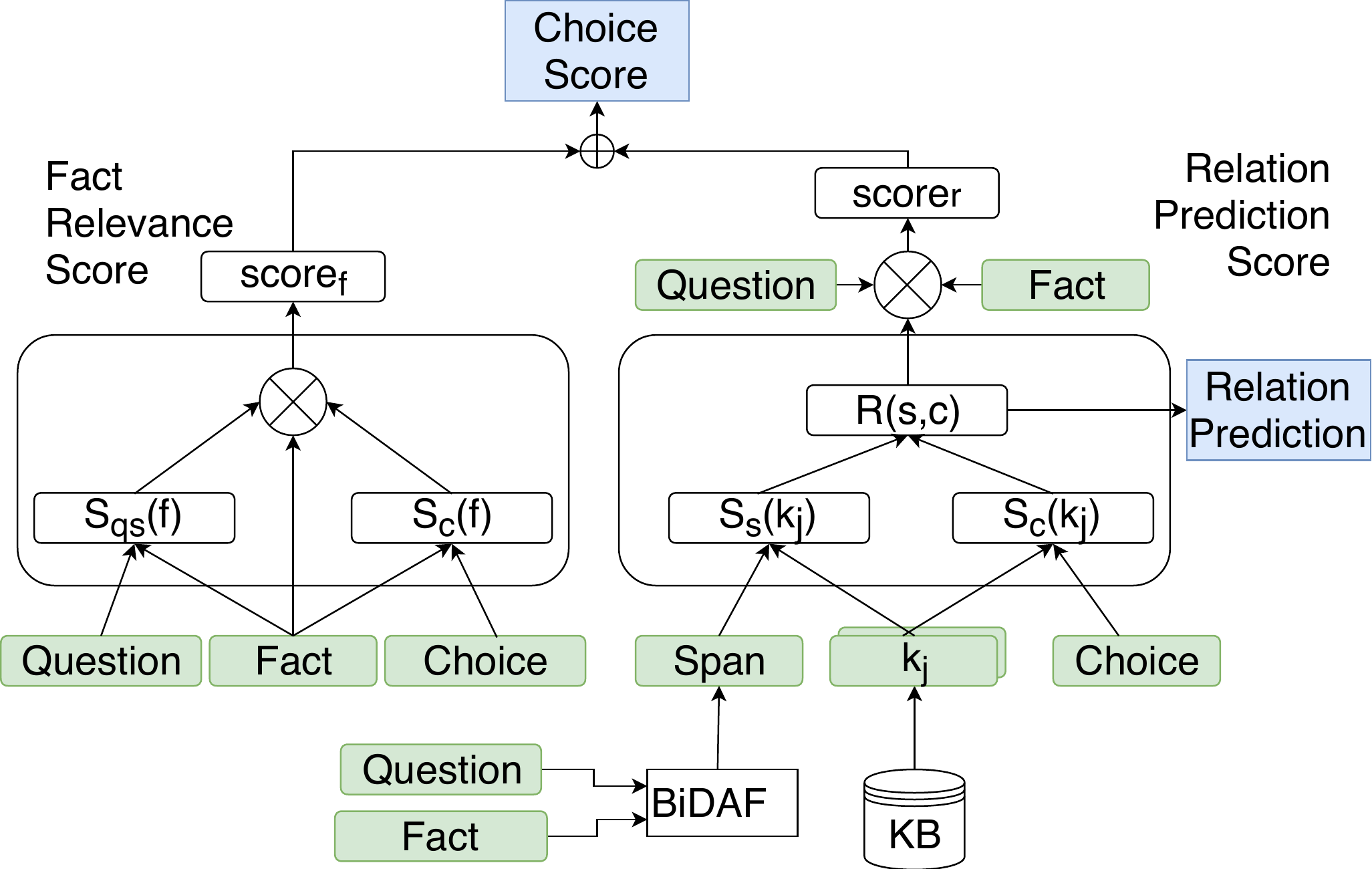}
    \caption{Overall architecture of the KGG question-answering model for each answer choice. The green nodes are the input to the model and the blue nodes are the model outputs that the losses are computed against. The model uses BiDAF to predict the key spans and retrieves KB facts based on the span and the input choice.}
    \label{fig:kg-model}
\end{figure}

The final score for each answer choice is computed by summing the fact relevance and relation prediction based scores i.e. $\score(\choicei) = \score_f(\choicei) + \score_r(\choicei)$. The final architecture of our QA model is shown in Figure~\ref{fig:kg-model}. 

\subsection{Model Training}

We use cross-entropy loss between the predicted answer scores $\pred{\choice}$ and the gold answer choice $\gold{\choice}$. Since we also have labels on the true relations between the gold span and the correct answer choice, we introduce an auxiliary loss to ensure the predicted relation $\relnrep$ corresponds to the true relation between $\spansym$ and $\choicei$. We use a single-layer feed-forward network to project $\relnrep(\spansym, \choicei)$ into a vector $\pred{\reln}_i \in \size{1}{l}$ where $l$ is the number of relations. Since multiple relations can be valid, we create an n-hot vector representation $\gold{\reln} \in \size{1}{l}$ where $\gold{\reln}[k]=1$ if $\reln_k$ is a valid relation. 

We use binary cross-entropy loss between the $\pred{\reln}_i$ and $\reln$ for the correct answer choice. For the incorrect answer choice, we do not know if any of the unselected relations(i.e. where $\reln[k]=0$) hold. But we do know that the relations selected by Turkers for the correct answer choice should not hold for the incorrect answer choice. To capture this, we compute the binary cross entropy loss between $\pred{\reln}_i$  and $1 - \reln$ for the incorrect answer choices but ignore the unselected relations.

Finally, the loss for each example, assuming \choicei is the correct answer, is given as $ \textrm{loss}  = ce(\pred{\choice}, \gold{\choices}) +  \lambda \cdot  \big(bce(\pred{\reln}_i, \gold{\reln}) + 
    \sum_{j \neq i} mbce(\pred{\reln}_j, 1 - \gold{\reln}, \gold{\reln})\big)$, where $ce$ is cross-entropy loss, $bce$ is binary cross-entropy loss, and $mbce$ is masked binary cross entropy loss, where unselected relations are masked.

We further augment the training data with questions in the \shortans dataset using the predicted spans and ignoring the relation loss. Also, we assume the labelled core fact in the OpenBookQA dataset provides the partial knowledge needed to answer these questions.

Implementation details and parameter settings are deferred to Appendix~\ref{app:impl}.  A sample visualization of the attentions and knowledge used in the model are provided in Figure~\ref{fig:model_att} in the Appendix.

\section{Experimental Results}

We present results of our proposed model, \gapqa, on two question sets: (a) those with short answers,\footnote{Answers with at most two non-stopword tokens.} \shortans (290 test questions), and (b) the complete set, \obqafull (500 test questions). As we mentioned before, \shortans subset is likely to have \ftoa gaps that can be targeted by our approach and we therefore expect larger and more meaningful gains on this subset. 

\subsection{Key Span Identification}

We begin by evaluating three training strategies for the key span identification model, using the annotated spans in \mfdataset for training. As seen in Table~\ref{tab:span_pred}, the BiDAF model trained on the SQuAD dataset~\cite{Rajpurkar2016-squad} performs poorly on our task, likely due to the different question style in OpenBookQA. While training on \mfdataset (from scratch) substantially improves accuracy, we observe that using \mfdataset to fine-tune BiDAF pre-trained on SQuAD results in the best F1 (78.55) and EM (63.99) scores on the Dev set. All subsequent experiments use this fine-tuned model. 

\begin{table}[t]
    \small
    \centering
    \setlength\extrarowheight{1pt}
    \setlength{\doublerulesep}{\arrayrulewidth}
    \begin{tabular}{l@{\hskip 5ex}cc}
        Training Data &  Dev F1 & Dev EM \\
        \hline\hline
        \T SQuAD &  54.67 & 41.40 \\
        \mfdataset & 72.99 & 58.60 \\
        \B SQuAD + \mfdataset(tuning) & {\bf 78.55} & {\bf 63.69}
    \end{tabular}
    \caption{BiDAF model performance on the span prediction task, under different choices of training data}
    \label{tab:span_pred}
\end{table}

\subsection{OpenBookQA Results}

We compare with three previous state-of-the-art \textbf{models} reported by \citet{openbookqa}. Two of these are Knowledge-free models (also referred to as No Context Baselines~\cite{chen-durrett-2019-understanding}): (a) Question-to-Choice (\textbf{Q2Choice}) computes attention between the question and the answer choice, (b) \textbf{ESIM + ELMo}, uses ESIM~\cite{esim-chen2017} with ELMo~\cite{elmo-Peters:2018} embeddings to compute question-choice entailment. The third is Knowledge Enhanced Reader (\textbf{KER}), which uses the core fact (f) and knowledge retrieved from ConceptNet to compute cross-attentions between the question, knowledge, and answer choices. 

For \textbf{knowledge}, we consider four sources: (1) \textbf{ConceptNet (CN)}, the English subset of ConceptNet v5.6.0 tuples;\footnote{https://github.com/commonsense/conceptnet5/wiki}  (2) \textbf{WordNet}, the WordNet subset of ConceptNet used by \citet{openbookqa}; (3) \textbf{OMCS}, the Open Mind Common Sense subset of ConceptNet used by \citet{openbookqa}; and (4) \textbf{ARC}, with 14M science-relevant sentences from the AI2 Reasoning Challenge dataset~\cite{ARCClark2018}.

Following OpenBookQA, we train each model five times using different random seeds, and report the average score and standard deviation (without Bessel's correction) on the test set. For simplicity and consistency with prior work, we report one std.~dev.\ from the mean using the $\mu \pm \sigma$ notation.

We train our model on the combined \mfdataset and \shortans question set with full supervision on examples in \mfdataset and only QA supervision (with predicted spans) on questions in \shortans. We train the baseline approaches on the entire question set as they have worse accuracies on both the sets when trained on the \shortans subset. We do not use our annotations for any of the test evaluations. We use the core fact provided by the original dataset and use the predicted spans from the fine-tuned BiDAF model. 

%
We present the test accuracies on the two question sets in Table~\ref{tab:results}. On the targeted \shortans subset, our proposed \gapqa improves statistically significantly over the partial knowledge baselines by 6.5\% to 14.4\%. Even though the full OpenBookQA dataset contains a wider variety of questions not targeted by \gapqa, we still see an improvement of 3+\% relative to prior approaches.

It is worth noting that recent large-scale language models (LMs)~\cite{bert,radford2018improving} have now been applied on this task, leading to improved state-of-the-art results~\cite{Sun2018ImprovingMR,banerjee-etal-2019-careful,pan2019improving}. However, our knowledge-gap guided approach to QA is orthogonal to the underlying model. Combining these new LMs with our approach is left to future work.

\begin{table}
    \small
    \setlength{\tabcolsep}{6pt}
    \setlength\extrarowheight{1pt}
    \centering
    \begin{tabular}{lll}
        Model & \shortans  & \obqafull\\
        \hline
        Q2Choice & 47.10 $\pm$ 1.5 & 49.64 $\pm$ 1.3 \\
        ESIM + ELMo & 45.93 $\pm$ 2.6 & 49.96 $\pm$ 2.5\\
        \kermodel (only f) & 57.93 $\pm$ 1.4  & 55.80 $\pm$ 1.8 \\
        \kermodel (f + WordNet) & 54.83 $\pm$ 2.5 & 55.84 $\pm$ 1.7 \\
        \kermodel (f + OMCS) & 49.65 $\pm$ 2.0 & 52.50 $\pm$ 0.8 \\
        \selectmod \gapqa (f + KB) [Ours]  & \textbf{64.41 $\pm$ 1.8}* &  \textbf{59.40 $\pm$ 1.3}* \\
    \end{tabular}
    \caption{Test accuracy on the the \shortans subset and \obqafull dataset assuming core fact is given.
    $*$ denotes the results are statistically significantly better than all the baselines (p$\leq$0.05, based on Wilson score intervals~\cite{wilson}).}
    \label{tab:results}
\end{table}

\paragraph{Effect of input knowledge.}
Since the baseline models use different knowledge sources as input, we evaluate the performance of our model using the same knowledge as the baselines.\footnote{The baselines do not scale to large scale corpora and so can not be evaluated against our knowledge sources.} Even when our model is given the same knowledge, we see an improvement by 5.9\% and 11.3\% given only WordNet and OMCS knowledge respectively. This shows that we can use the available knowledge, even if limited, more effectively than previous methods. When provided with the full ConceptNet knowledge and large-scale text corpora, our model is able to exploit this additional knowledge and improve further by 4\%. 
\begin{table}[t]
   \small
    \centering
    \setlength{\tabcolsep}{12pt}
    \setlength{\doublerulesep}{\arrayrulewidth}
    \begin{tabular}{lll}
        Knowledge Source & Model & \shortans \\
        \hline\hline
        \T \multirow{ 2}{*}{f + WordNet} & \kermodel  & 54.83 $\pm$ 2.5 \\
        & \gapqa & \textbf{60.69 $\pm$ 1.1}* \\
        \hline
        \T \multirow{ 2}{*}{f + OMCS} & \kermodel  & 49.65 $\pm$ 2.0 \\
        & \gapqa  & \textbf{60.90 $\pm$ 2.4}* \\
        \hline
        \T f + CN + ARC & \gapqa & \textbf{64.41 $\pm$ 1.8} \\
    \end{tabular}
    \caption{Test accuracy on the \shortans subset with different sources of knowledge. $*$ denotes the results are statistically significantly better than the corresponding KER result (p$\leq$0.05, based on Wilson score intervals~\cite{wilson}).
    }
    \label{tab:full_set}
\end{table}

\subsection{Ablations}

\begin{table}[t]
    \small
    \setlength\extrarowheight{1pt}
    \centering
    \setlength{\doublerulesep}{\arrayrulewidth}
    \begin{tabular}{l@{\hskip 8ex}cc}
        Model & \shortans & $\Delta$ \\
        \hline
        \hline
        \T \gapqa   & \textbf{64.41 $\pm$ 1.8} & ---\\
        \; No Annotations &  58.90 $\pm$ 1.9 & 5.51 \\
        \; Heuristic Span Anns. &  61.38 $\pm$ 1.5 & 3.03 \\ 
        \; No Relation Score & 60.48 $\pm$ 1.1 & 3.93 \\
        \; No Spans (Model) &  62.14 $\pm$ 2.1 & 2.27 \\
        \B \; No Spans (IR) &  61.79 $\pm$ 1.0 & 2.62
    \end{tabular}
    \caption{Average accuracy of various ablations, showing that each component of \gapqa has an important role. No Annotations = Ignore Span \& Relation Annotations, Heuristic Span Anns = Heuristically predict span annotations (no human annotations), No Relation Score = Ignore the relation-based score ($score_r$), No Spans (Model) = Ignore the span (use entire fact) to compute span-weighted representations, No Spans (IR) = Ignore the span (use entire fact) for retrieval.
    }
    \label{tab:ablation}
\end{table}

\begin{table*}[ht]
    \setlength{\doublerulesep}{\arrayrulewidth}
    \centering
    \small
    \begin{tabular}{p{5cm}|p{3.2cm}|p{1.2cm}|p{4.8cm}}
        \multirow{2}{*}{Question} & \multirow{2}{*}{Fact} & Predicted Answer & \multirow{2}{*}{Reason} \\
        \hline\hline
        \T What vehicle would you use to travel on the majority of the surface of the planet on which we live? (A) Bike (B) \textbf{Boat} (C) Train (D) Car & oceans cover 70\% of the surface of the earth &  Bike & Predicted the \textbf{wrong span} ``70\%''. \\
        \hline
        \T What contains moons? (A) ships (B)\textbf{space mass} (C) people (D) plants & the solar system contains the moon & ships & \textbf{Scores the relation for the incorrect answer higher} because of the facts connecting ``systems'' and ``ships''. \\
        \hline
        \T Cocoon creation occurs (A) \textbf{after the caterpillar stage} (B) after the chrysalis stage (C) after the eggs are laid (D) after the cocoon emerging stage & the cocoons being created occurs during the pupa stage in a life cycle &  after the chrysalis stage & Does not model the \textbf{complex relation} (temporal ordering) between the key span: ``pupa stage'' and  ``caterpillar stage''. Instead it predicts ``chrysalis'' due to the synonymy with ``pupa''.
    \end{tabular}
    \caption{Sample errors made by the \gapqa on questions from the \shortans dataset. The correct answers are marked in \textbf{bold} within the question.}
    \label{tab:err_egs}
\end{table*}

We next evaluate key aspects of our model in an ablation study, with average accuracies in Table~\ref{tab:ablation}.

\textbf{No Annotations (No Anns)}: 
We ignore all collected annotations (span, relation, and fact) for training the model. We use the BiDAF(SQuAD) model for span prediction, and only the question answering loss for the QA model trained on the \shortans subset.\footnote{Model still predicts the latent relation representation.} Due to the noisy spans produced by the out-of-domain BiDAF model, this model performs worse that the full \gapqa model by 5.5\% (comparable performance to the \kermodel models). This shows that our model does utilize the human annotations to improve on this task.

\textbf{Heuristic Span Annotations}: We next ask whether some of the above loss in accuracy can be recovered by heuristically producing the spans for training---a cost-effective alternative to human annotations. We find the longest subsequence of tokens (ignoring stop words) in \fact that is not mentioned in \ques and assume this span (including the intermediate stop words) to be the key term. To prevent noisy key terms, we only consider a subset of questions where 60\% of the non-stopword stemmed tokens in \fact are covered by \ques. We fine-tune the BiDAF(SQuAD) model on this subset and then use it to predict the spans on the full set.\footnote{We use the questions from the \mfdataset + \shortans set. Note we are only evaluating the impact of heuristic spans compared to human-authored spans, but assume that we have good quality partial context as provided in \mfdataset.} We train \gapqa model on this dataset without any relation labels (and associated loss). This simple heuristic leads to a 3\% drop compared to human annotations, but still out-performs previous approaches on this dataset, showing the value of the gap-based QA approach.

\textbf{No Relation Score}: We ignore the entire relation-based score ($score_r$) in the model and only rely on the fact-relevance score. The drop in score by 3.9\% shows that the fact alone is not sufficient to answer the question using our model.

\textbf{No Spans (Model)}: We ignore the spans in the model, i.e., we use the entire fact to compute the span-based representation $ \wtrep{\pred{\spansym}}{\kbj}$. In effect, the model is predicting the gap between the entire fact and answer choice.\footnote{Retrieval is still based on the span and we ignore the relation prediction loss.} We see a drop of $\sim$2\%, showing the value of spans for gap prediction.

\textbf{No Spans (IR)}: Ignoring the span for retrieval, the knowledge is retrieved based on the entire fact (full \gapqa model is used). The drop in accuracy by 2.6\% shows the value of targeted knowledge-gap based retrieval.

\subsection{Error Analysis}
\label{app:error}

We further analyzed the performance of \gapqa on 40 incorrectly answered questions from the dev set in the \shortans dataset. Table~\ref{tab:err_egs} shows a few error examples. There were three main classes of errors:\\
    \noindent \textbf{Incorrect predicted spans} (25\%) often due to complex language in the fact or the \qtof gap needed to accurately identify the span.
    
    \noindent \textbf{Incorrect relation scores} (55\%) due to distracting facts for the incorrect answer or not finding relevant facts for the correct answer, leading to an incorrect answer scoring higher.
    
    \noindent \textbf{Out-of-scope gap relations} (20\%) where the knowledge gap relations are not handled by our model such as temporal relations or negations (e.g., is \emph{not} made of).

Future work in expanding the dataset, incorporating additional relations, and better retrieval could mitigate these errors.

\section{Conclusion}

We focus on the task of question answering under partial knowledge: a novel task that lies in-between open-domain QA and reading comprehension. We identify classes of knowledge gaps when reasoning under partial knowledge and collect a dataset targeting one common class of knowledge gaps. We demonstrate that identifying the knowledge gap first and then reasoning by filling this gap outperforms previous approaches on the OpenBookQA task, with and even without additional missing fact annotation. This work opens up the possibility of focusing on other kinds of knowledge gaps and extending this approach to other datasets and tasks (e.g., span prediction).

\section*{Acknowledgements}
We thank Amazon Mechanical Turk workers for their help with annotation, and the reviewers for their invaluable feedback. Computations on beaker.org were supported in part by credits from Google Cloud.

\bibliographystyle{acl_natbib}
\bibliography{knowledge_gap}

\clearpage
\appendix

\section{Annotating Gaps: Data Collection}
\label{appendix:data-collection}
\noindent
We first identify relevant facts for questions and then collect annotations for fact-answer gap, given the relevant fact. However, straightforward approaches to annotate all questions led to noisy labels. To improve annotation quality, we identified question subsets most suitable for this task and split the fact-answer gap annotation into two steps.

\paragraph{Fact Relevance.}
The OpenBookQA dataset provides the core science fact used to create the question. However, in 20\% of the cases, while the core science fact inspired the question, it is not needed to answer the question~\cite{openbookqa}. We also noticed that often multiple facts from the open book can be relevant for a question. So we first create an annotation task to identify the relevant facts from a set of retrieved facts. Also to ensure that there is a gap between the fact and the correct answer, we select facts that have no word overlap with the correct choice or have overlap with multiple answer choices. This ensures that the fact alone can not be trivially used to answer the question.

We ask Turkers to annotate these retrieved facts as (1) are they \emph{relevant} to the question and (2) if relevant, do they point to a \emph{unique} answer. We introduced the second category after noticing that some generic facts can be relevant but not point to a specific answer making identifying the knowledge gap impossible. E.g. The fact: ``evaporation is a stage in the water cycle process'' only eliminates one answer option from ``The only stage of the water cycle process that is nonexistent is (A) evaporation (B) evaluation (C) precipitation (D) condensation''. For each question, we selected facts that were marked as relevant and unique by at least two out of three turkers. 

\paragraph{Knowledge Gap.}
In the second round of data collection, we asked Turkers to write the facts connecting the relevant fact to the correct answer choice. We restricted this task to Masters level Turkers with 95\% approval rating and 5000 approved hits. However, we noticed that crowd-source workers would often re-state part of the knowledge mentioned in the original fact or directly connect the question to the answer. This issue was also mentioned by the authors of OpenBookQA who also noticed that the additional facts were ''\emph{noisy (incomplete, over-complete, or only distantly related)}''~\cite{openbookqa}. E.g. for the question: ``In the desert, a hawk may enjoy an occasional (A) coyote (B) reptile (C) bat (D) scorpion`` and core fact: ``hawks eat lizards'', one of the turk-authored additional fact: ``Hawks hunt reptiles which live in the desert'' is sufficient to answer the question on its own.

We also noticed that questions with long answer choices often have multiple fact-answer gaps leading to  complex annotations, e.g. ``tracking time'' \emph{helps with} ``measuring how many marshmallows I can eat in 10 minutes''. 
 Collecting knowledge gaps for such questions and common-sense knowledge to capture these gaps are interesting directions of future research. We instead focus on questions where the answer choices have at most two non-stopword tokens. We refer to this subset of questions in OpenBookQA as \emph{\shortans}, which still forms more than 50\% of the OpenBookQA set. This subset also forms the target question set of our approach.

 Further to simplify this task, we broke the task of identifying the required knowledge into two steps (shown in Figure~\ref{fig:turk_task} in Appendix): (1) identify key terms in the core fact that could answer the question, and (2) identify the relationship between these terms and the correct answer choice.  For key terms, we asked the Turkers to select spans from the core fact itself, to the extent possible. For the relation identification, we provided a list of relations and asked them to select all the relations that hold between the key term and the correct choice but do not hold for the incorrect answer choices. Based on our analysis, we picked nine most common relations: \{causes, definedAs, enables, isa, located in, made of, part of, provides, synonym of\} and their inverses (except synonymy).\footnote{These relations were also found to be important by prior approaches~\cite{clark2014automatic,tableilp2016,Jansen2016Explanation,Jansen2018WorldTreeAC} in the science domain.} If none of these relations were valid, they were allowed to enter the relation in a text box.

We note that the goal of this effort was to collect supervision for a subset of questions to guide the model and show the value of minimal annotation on this task. We believe our approach can be useful to collect annotations on other question sets as well, or can be used to create a challenge dataset for this sub-task. Moreover, the process of collecting this data revealed potential issues with collecting annotations for knowledge gaps and also inspired the design of our two-step QA model. 

\begin{figure*}
    \centering
    \includegraphics[scale=0.48, trim={0.5cm 0 9.5cm 0}, clip]{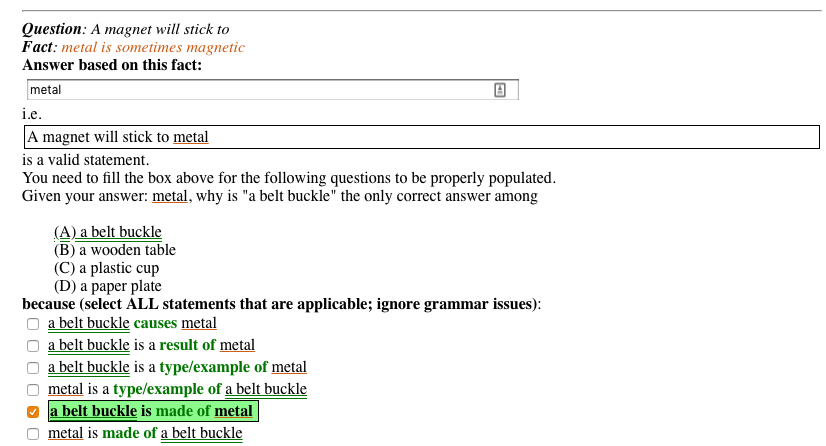}
    \caption{Interface provided to Turkers to annotate the missing fact. Entering the answer span from the fact, metal, in this example, automatically populates the interface with appropriate statements. The valid statements are selected by Turkers and capture the knowledge gap.}
    \label{fig:turk_task}
\end{figure*}

\begin{figure*}
    \centering
    \includegraphics[scale=0.4]{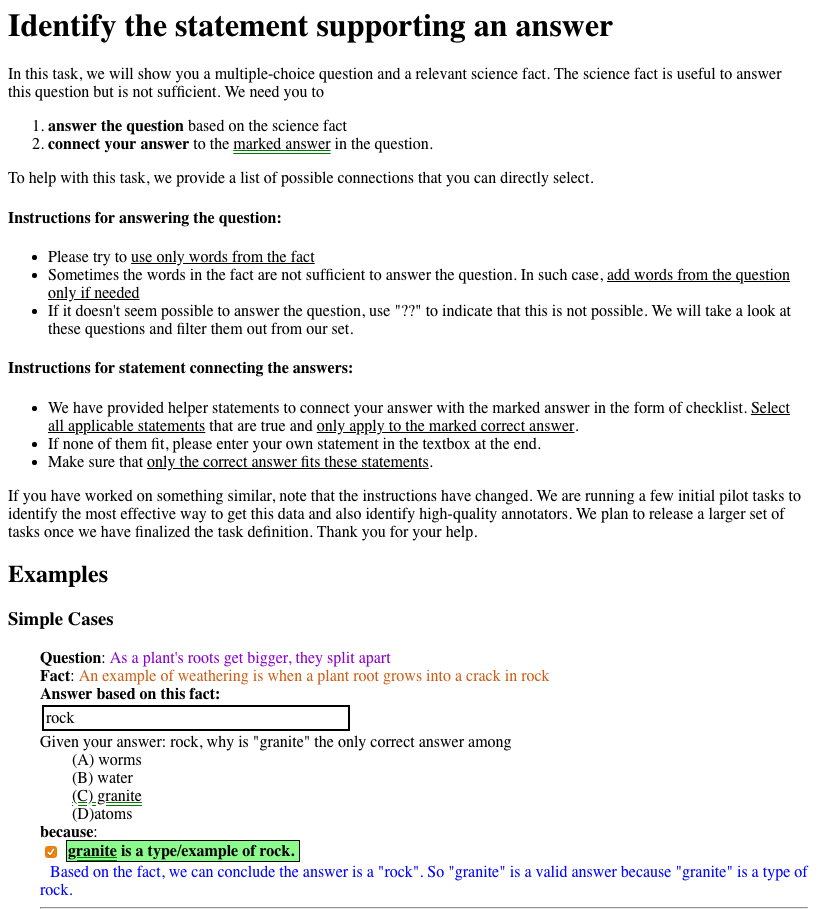}
    \caption{Basic Instructions for the task}
    \label{fig:basic_instr}
\end{figure*}

\begin{figure*}
    \centering
    \includegraphics[scale=0.44]{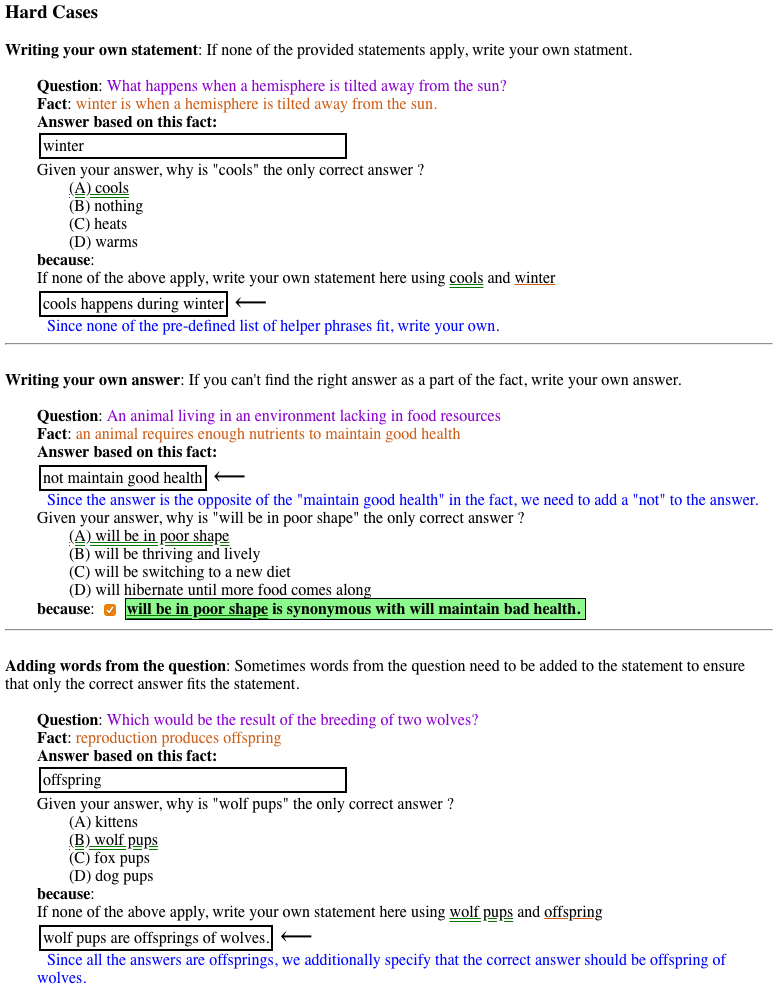}
    \caption{Instructions for complex examples}
    \label{fig:hard_instr}
\end{figure*}

\begin{figure*}
 \centering
     \includegraphics[width=\linewidth]{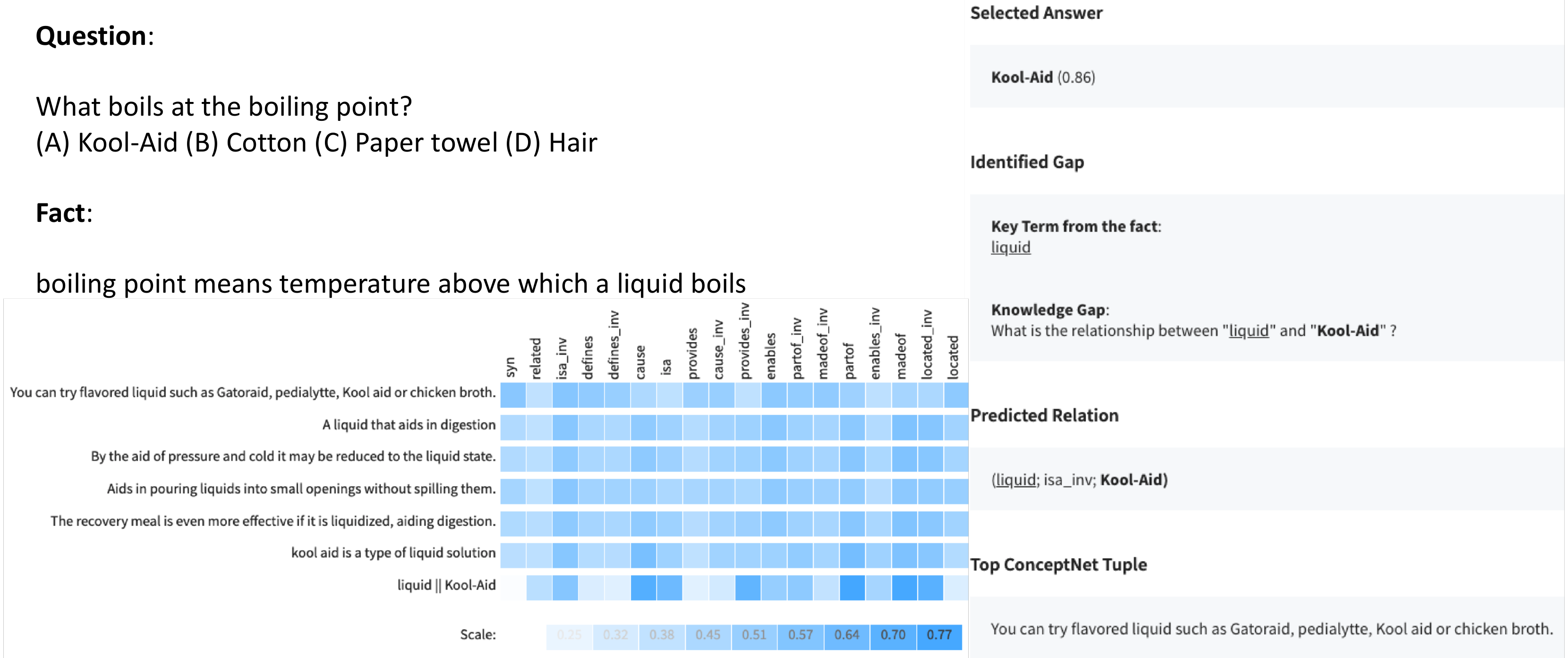}
     \caption{Visualization of the models behavior with the predicted span, top predicted relation, and the top fact used by model. The heat map shows the confidence of the model for all the relations for each input sentence (first five) and ConceptNet sentencized tuple (last but one) and the back-off tuple (last one) to capture the knowledge in the embeddings.}
     \label{fig:model_att}
\end{figure*}
 
\section{Implementation Details}
\label{app:impl}
We implement all our models in Pytorch~\cite{paszke2017automatic} using the AllenNLP~\cite{Gardner2017AllenNLP} toolkit. We also used the AllenNLP implementation of the BiDAF model for span prediction. We use 300D 840B Glove~\cite{pennington2014glove} embeddings and use 200 dimensional hidden representations for the BiLSTM shared between all inputs (each direction uses 100 dimensional hidden vectors). We use 100 dimensional representations for the relation prediction, $\relnrep_j$. Each feedforward network, $\mlp$ is a 2-layer network with relu activation, 0.5 dropout~\cite{dropout}, 200 hidden dimensions on the first layer and no dropout on the output layer with linear activation. We use a variational dropout~\cite{Gal2016DropoutAA} of 0.2 in all the BiLSTMs. The relation prediction loss is scaled by $\lambda=1$. We used the Adam~\cite{adam} optimization with initial $lr=0.001$ and a learning rate scheduler that halves the learning rate after 5 epochs of no change in QA accuracy. We tuned the hyper-parameters and performed early stopping based on question answering accuracy on the validation set. Specifically, we considered \{50, 100, 200\} dimensional representations, $\lambda \in \{0.1, 1, 10\}$, retrieving \{10, 20\} knowledge tuples and \{[x - y; x*y], [x, y]\} combination functions for $\bigotimes$ during the development of the model. The baseline models were developed for this dataset using hyper-parameter tuning; we do not perform any additional tuning. Our model code and pre-trained models are available at https://github.com/allenai/missing-fact.

\section{ConceptNet sentences}
\label{sec:conceptnet}
 Given a tuple $t=(s, v, o)$, the sentence form is generated as ``$s$ is $\mathrm{split}(v)$ $o$'' where $\mathrm{split}(v)$ splits the ConceptNet relation $v$ into a phrase based on its camel-case notation. For example, (belt buckle, /r/MadeOf, metal) would be converted into ``belt buckle is made of metal''.
 
 \section{Text retrieval}
 \label{app:text_retrieval}
 For each span \pred{\spansym} and answer choice \choicei, we query an ElasticSearch~\footnote{https://www.elastic.co/} index on the input text corpus with the ``\pred{\spansym} + \choicei'' as the query. We also require the matched sentence must contain both the span and the answer choice. We filter long sentences ($>$300 characters), sentences with negation and noisy sentences\footnote{Sentences are considered clean if they contain alpha-numeric characters with standard punctuation, start with an alphabet or a number, are single sentence and only uses hyphens in hyphenated word pairs} from the retrieved sentences.

\end{document}